\documentclass[10pt, letterpaper, twocolumn]{article}
\usepackage{dsperse_whitepaper}

\newcommand{\projectName}{DSperse}

\newcommand{\numDataPoints}{70}

\title{{\projectName}: A Framework for Targeted Verification in Zero-Knowledge Machine Learning}

\author{\vspace{1em}
  {\normalsize Dan Ivanov,} {\normalsize Tristan Freiberg,} {\normalsize Shirin Shahabi,} {\normalsize Jonathan Gold, and} {\normalsize Haruna Isah}\\
  {\normalsize [\href{mailto:dan@inferencelabs.com}{dan,} \href{mailto:tristan@inferencelabs.com}{tristan,} \href{mailto:shirin@inferencelabs.com}{shirin,} \href{mailto:jonathan@inferencelabs.com}{jonathan, and} \href{mailto:haruna@inferencelabs.com}{haruna}]}@inferencelabs.com\\
  {\normalsize Inference Labs Inc.} 
}

\date{\today}

\begin{document}

\twocolumn[
\begin{@twocolumnfalse}
\maketitle
\begin{abstract}
\noindent
{\projectName} is a modular framework for distributed machine learning inference with strategic cryptographic verification. Operating within the emerging paradigm of \emph{distributed zero-knowledge machine learning}, {\projectName} avoids the high cost and rigidity of full-model circuitization by enabling \emph{targeted verification} of strategically chosen subcomputations. These verifiable segments, or ``slices'', may cover part or all of the inference pipeline, with global consistency enforced through audit, replication, or economic incentives. This architecture supports a pragmatic form of \emph{trust minimization}, localizing zero-knowledge proofs to the components where they provide the greatest value. We evaluate {\projectName} using multiple proving systems and report empirical results on memory usage, runtime, and circuit behavior under sliced and unsliced configurations. By allowing proof boundaries to align flexibly with the model's logical structure, {\projectName} supports scalable, targeted verification strategies suited to diverse deployment needs.
\end{abstract}
\tableofcontents
\vspace{1em}

\end{@twocolumnfalse}
]

\section{Introduction}

As AI models are increasingly deployed in decentralized environments, ranging from distributed compute networks to on-chain inference, the need for verifiable machine learning has become more urgent. This has led to growing interest in the emerging space of \emph{distributed zero-knowledge machine learning} (dzkML), which combines cryptographic proofs with modular, multi-agent inference pipelines. In this setting, inference may be distributed across untrusted nodes, and model providers require guarantees that critical subcomputations are executed faithfully, without revealing proprietary weights or internal logic \cite{ghodsi2017safetynets}.

Zero-knowledge proofs (ZKPs) offer a compelling tool for this purpose, but current approaches remain too computationally expensive to scale to real-world models \cite{peng2025survey, sheybani2025zero}. These inefficiencies arise from the large arithmetic circuits and proof objects required for zkML, whose size and computational cost scale with the complexity of the model. Full-model circuitization, required for end-to-end cryptographic guarantees, introduces prohibitive cost and latency, rendering zkML infeasible for many practical deployments, especially in machine learning-as-a-service (MLaaS) contexts \cite{south2024verifiable}.

{\projectName} addresses this bottleneck by enabling \emph{targeted verification}: a slice-based architecture in which only strategically selected segments of a model are circuitized and proven. These segments may include proprietary logic, safety-critical routines, or other high-value computations. In many real-world systems, selectively verifying high-leverage components may offer a more practical and scalable alternative to full-model verification, particularly in complex or distributed settings. 

For example, in a financial fraud detection pipeline, fully circuitizing and verifying a large, frequently updated model may be infeasible, especially if retraining changes the circuit structure or requires costly recompilation. However, selectively proving the execution of key components, such as anomaly detection modules or decisions trees, can provide useful assurances about the integrity of critical subcomputations. This approach enables developers to reduce proving cost and latency while preserving verifiability where it matters most. Slices can be independently verified and flexibly composed, offering a modular strategy for balancing trust, performance, and deployability.

Our results suggest that distributed inference and zkML frameworks like {\projectName} can offer meaningful verifiability over selected subcomputations, without incurring the full cost of monolithic proofs. This makes them well-suited for deployment in real-world systems where partial verifiability offers practical benefits, even if full inference guarantees are out of reach.

\section{Related Work}

The zkML landscape is comprised of verifiable training, testing, and inference. As outlined by Peng et al.\ \cite{peng2025survey}, these collectively ensure trust in machine learning by confirming that training meets client-specified data and model requirements, testing accurately reflects the model's generalization ability, and inference produces correct predictions using the designated model and process, while preserving confidentiality. Although there is a growing body of literature on verifiable training and testing, this work focuses on inference, arguably the most exposed and latency-sensitive phase in many real-world ML deployments. Inference is also where ZKP-based assurances are currently most feasible, and where they may offer the greatest near-term impact \cite{scaramuzza2025engineering}.

A key limitation of verifiable ML inference with ZKPs, especially for large or complex models, is the substantial computational overhead and slow proof generation \cite{south2024verifiable}. Most zkML research to date has focused on full-model circuitization to ensure end-to-end cryptographic guarantees. However, this incurs prohibitive cost and latency, limiting practical deployment. According to a recent survey by Xing et al.\ \cite{xing2025zero}, efficiency gains can be pursued through three main avenues: (i) tailoring proof system designs to specific ML models or exploring alternative ZKP systems beyond Quadratic Arithmetic Program (QAP)-based solutions; (ii) leveraging specialized hardware such as Field-Programmable Gate Arrays (FPGAs), Graphics Processing Units (GPUs), and pipelined accelerators; and (iii) balancing security and privacy with efficiency. Our study adopts option (iii), which allows for selective or targeted verification and is key to the practical applicability of zkML.

One recent system, psvCNN \cite{fan2024psvcnn}, addresses the challenge of high proof burdens for full-model CNN inference by parallelizing the circuit into computationally independent blocks. This enables efficient proof generation across multiple cores or distributed nodes and demonstrates substantial speedups over previous zkML approaches. However, psvCNN maintains an end-to-end proof model, which may still be prohibitive for frequent or large-scale inference in constrained environments.

{\projectName} takes a different approach: rather than proving the entire inference process, it supports modular and selective verification of strategically chosen segments. This allows developers to reduce resource demands by focusing on the highest-value computations. While this sacrifices the global guarantees of a full proof, it offers a more scalable and flexible tradeoff for scenarios where targeted trust is sufficient and computational resources are limited.

Model slicing, the technique of segmenting parts of a deep neural network, is not entirely a new concept, having been explored in prior work by Zhang et al.\ \cite{zhang2020dynamic} and Zhou et al.\ \cite{zhou2024neusemslice}. However, {\projectName} applies this idea in a novel cryptographic setting. Our contribution lies in a pragmatic framework for selectively verifying high-value subcomputations during inference, using independently provable slices. This targeted approach enables more efficient and flexible verification pipelines. To our knowledge, slicing strategies have not previously been applied to zkML inference.

\section{{\projectName} Overview}

\subsection{System Goals}

{\projectName} is a pragmatic framework for deploying machine learning models in decentralized environments where full zero-knowledge inference remains, for now, prohibitively expensive. For most models of practical interest, circuitizing the entire computation introduces unacceptable overhead in terms of latency, proving cost, and fidelity loss. In many real-world scenarios, however, components of a model vary in their criticality, and so too may their verification requirements. For example, in a self-driving car system, the submodel responsible for obstacle detection or emergency maneuvers may warrant end-to-end cryptographic verification to ensure safety and accountability. In contrast, auxiliary models that handle environmental monitoring or route suggestions could be selectively verified, or monitored via other mechanisms, to reduce overhead without compromising essential guarantees. {\projectName} supports such a risk-sensitive approach to verification, offering developers a way to focus resources where they matter most while maintaining a modular architecture compatible with evolving trust frameworks.

Rather than aiming for universal cryptographic enforcement, {\projectName} focuses on what can be verified efficiently and usefully in practice. It delivers meaningful assurances precisely where they are needed, over the most sensitive and proprietary parts of a model, while avoiding the performance penalties associated with full-circuit approaches. {\projectName} enables developers to isolate and verify high-value segments, thereby improving trust, auditability, and deployment feasibility of ML models without imposing unrealistic constraints. In this way, it offers a practical and scalable approach to incorporating verifiability into modern ML pipelines, aligned with the real-world demands of infrastructure and use cases.

At the same time, {\projectName} is designed with a long-term goal in mind: to serve as a foundation for future systems that support end-to-end cryptographic verification. Its architecture anticipates modular proof composition, recursive linking, and more advanced forms of integrity enforcement. 

\subsection{Capabilities of {\projectName}}

{\projectName} provides a flexible and modular framework for distributed ML inference, with optional cryptographic verification of selected components. Its primary goal is to enable fine-grained control over how a model is ``sliced'' into subcomputations, which can then be executed, circuitized, or verified independently. These slices may span part or all of the inference pipeline: the system places no upper bound on coverage. While each segment is proven independently, linking those segments into a single proof boundary is deferred to higher-level orchestration. This decomposition allows developers to strike a practical balance between verifiability, performance, model confidentiality, and resource constraints.

In terms of verifiability, {\projectName} allows model providers to selectively circuitize parts of a neural network, such as final classification layers or other proprietary modules, and to generate ZKPs of correct execution for those segments. The remaining parts of the model, often standard architectures or publicly available components, can be run openly by the user or delegated to public infrastructure. By minimizing the scope of what must be circuitized, {\projectName} reduces proving overhead and preserves fidelity (see \ref{sec:fidelity}) relative to the original floating-point model.

{\projectName} also supports purely decentralized inference without cryptographic proof, by distributing model execution across a network of compute nodes. Even in this non-verifiable mode, the system retains utility as a lightweight, scalable method for inference delegation, with optional audit logging or reproducibility mechanisms. In both modes, {\projectName} offers granular control over computational workload and memory usage. The architecture exposes parameters that allow circuit designers to specify how many layers of a model are included in each slice, enabling adaptation to the capabilities of resource-constrained nodes. This is particularly useful in distributed environments, where RAM and compute limitations vary significantly across devices.

While {\projectName} does not currently provide automated proof composition across slices, it places no restrictions on full-model verification. A model provider may choose to circuitize the entire inference as a single slice, or verify each slice independently with output-to-input consistency managed externally. This flexibility allows {\projectName} to accommodate deployments ranging from partial verification of proprietary components to full-model coverage, depending on the specific use case's trust model and orchestration logic, without requiring changes to the system's core design.

In short, {\projectName} is designed not as a rigid proving system but as an adaptable foundation for decentralized, selectively verifiable inference. Its focus is on giving developers meaningful control over the structure, visibility, and verifiability of model components, with the goal of enabling real-world deployment of cryptographically grounded ML services under practical constraints.

\subsection{High-Level Architecture}

\begin{figure}[ht]
\centering
\includegraphics[width=0.9\columnwidth]{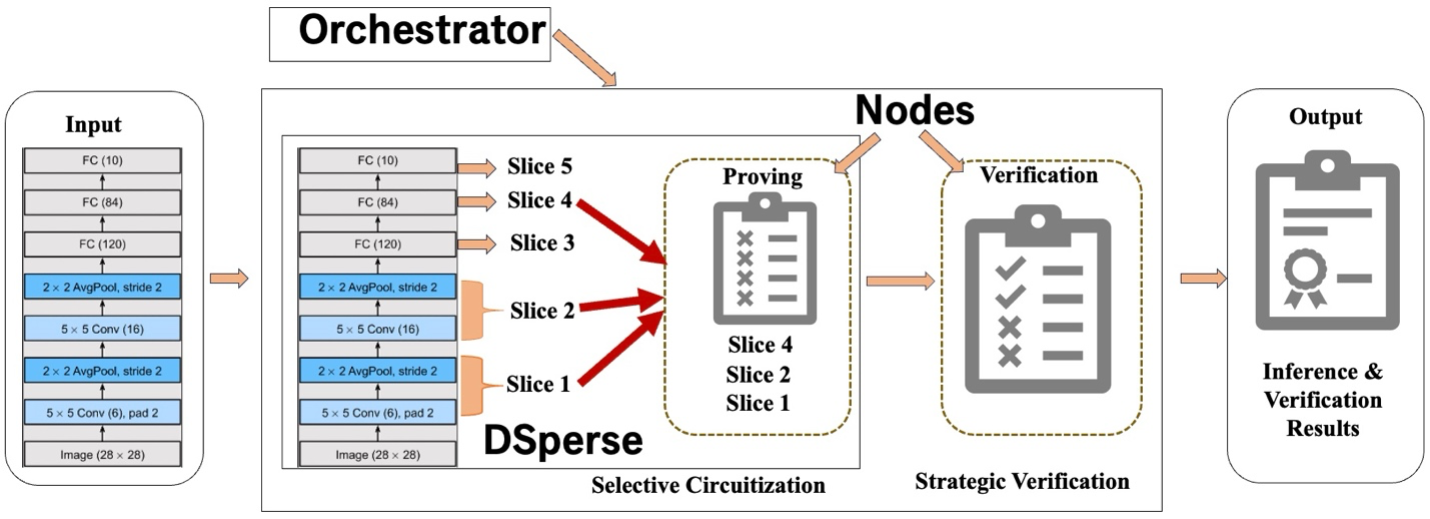}
\caption{DSperse framework architecture}
\label{fig:high_level_architecture}
\end{figure}

In its current form, {\projectName} is best understood as a conceptual framework for distributing and selectively verifying segments of an ML inference pipeline, rather than a fully specified dzkML protocol. The user submits a model and inference input data. As shown in Figure \ref{fig:high_level_architecture}, {\projectName} provides tools for model slicing, circuit generation, and per-slice proof execution, but leaves system-level concerns, such as key management, consistency enforcement, and result aggregation, to external infrastructure or an Orchestrator.

The Orchestrator divides the model into sequentially dependent slices (i.e., contiguous subsets of layers), assigns each slice to a Node, and coordinates the flow of intermediate values between Nodes. Each Node processes its assigned slice by executing the computation, generating the corresponding witness, and producing a ZKP of correct execution.

The system does not enforce global soundness cryptographically. Instead, correctness and consistency across slices must be ensured via additional mechanisms such as audit, redundancy, or external verification protocols. This modular approach allows for flexible deployment across heterogeneous environments and enables partial cryptographic assurances over performance-critical or trust-sensitive parts of the model, without incurring the overhead of full-model circuitization.

The final output includes the model's prediction and a collection of per-slice proofs, which can be individually verified to confirm that certain computations were performed correctly, without revealing proprietary model details or user data.

\FloatBarrier
\FloatBarrier

The design localizes ZKPs to only the most critical slices of the model, reducing proving cost, while allowing the remaining layers to execute without proof. The flow of operations is as follows:

\begin{itemize}[noitemsep, topsep=0pt]
    \item User (Input): submits the ML model (weights/parameters) and inference data.
    \item Slicing Module: the DSperse framework splits the model into discrete slices (sub-networks), each of which can be independently proven, with respect to its local computation.
    \item Prover Node: receives a model slice and its input (which may be an intermediate activation), computes the slice's output, and passes the results to the proof generator.
    \item Proof Generation Module: wraps the slice's computation in a ZKP of correct execution (generating, for example, an EZKL \cite{ezkl2025} proof).
    \item Verifier Node: validates each ZKP. These proofs confirm that specific subcomputations were correctly executed. Additional trust assumptions or mechanisms are required to ensure correctness of the full inference pipeline.
    \item Caching/Batching: the system may cache slice outputs or batch inputs and dynamically assign layer ranges to different nodes to improve efficiency.
    \item Output: component that stores the model's final prediction along with the corresponding per-slice proof artifacts and verification outcomes.
\end{itemize} 

\subsection{Architecture Constraints}

{\projectName} is designed to support partial verification of ML models by splitting inference into discrete, independently verifiable subcomputations. To enable this, models must conform to certain architectural constraints. First, the system assumes a unidirectional dataflow in which model inference proceeds through a fixed sequence of layers or modules. Each segment receives its inputs, performs a computation, and outputs its results without backward connections, dynamic control flow, or reuse of intermediate tensors across segments. This structure aligns with standard feedforward neural networks, including convolutional and multilayer perceptrons, as well as many transformer variants during inference. {\projectName}-compatible models can be viewed as computation graphs with a directed acyclic structure, where execution proceeds along a topological ordering of subcomputations. Each such subcomputation may be circuitized and verified independently, provided it respects locality and parameter isolation.

Architectures with loops or dynamic iteration, such as recurrent neural networks (RNNs) or long short-term memory (LSTMs), are currently unsupported. Similarly, attention mechanisms that depend on access to intermediate activations across circuit boundaries are not feasible unless the entire attention block is contained within a single circuitized segment.

To ensure that each circuitized segment is self-contained and verifiable in isolation, {\projectName} also disallows parameter reuse across slices. Parameters used in one segment must not be accessed by another unless the reuse occurs entirely within the same slice, with no dependency spanning circuit boundaries.

These constraints imply that slicing strategies must follow natural layer boundaries and avoid fragmenting operations with internal dependencies or shared state. While this limits the class of models {\projectName} supports out of the box, it encompasses many practical architectures, including those used in image classification, tabular inference, or transfer learning scenarios where base layers are public and task-specific heads are proprietary.

{\projectName} is intended as a foundation for building verifiable inference pipelines, not as a general-purpose verifier for arbitrary computational graphs. Future versions may explore broader support for recurrence, branching, or shared-state execution through compositional proof techniques or recursive circuit synthesis.

\subsection{System Guarantees}

{\projectName} provides a set of guarantees that reflect its pragmatic design. It offers \emph{targeted verifiability}: only strategically selected segments of a computation are circuitized and cryptographically proven. These segments typically contain proprietary logic, sensitive parameters, or otherwise high-value components. The remainder of the inference pipeline proceeds without formal cryptographic guarantees, but may still be subject to audit, replication, or economic incentive mechanisms. This architecture enables a form of \emph{trust minimization}. Rather than requiring users to trust the entire inference process, {\projectName} reduces the trust surface to only those parts that lie outside the verifiable scope. Trust is not eliminated, but it is explicitly localized and, when possible, replaceable.

{\projectName} supports \emph{strategic circuitization}, allowing developers to decide which subcomputations merit formal verification. Conceptually, an inference can be decomposed as:
\[
z = F_{\text{pub}}^{(2)} \circ F_{\text{priv}} \circ F_{\text{pub}}^{(1)}(x),
\]
where $F_{\text{pub}}^{(1)}$ and $F_{\text{pub}}^{(2)}$ are public computations (e.g., preprocessing and postprocessing), and $F_{\text{priv}}$ is a sensitive intermediate function that is circuitized and proven in zero knowledge. {\projectName} certifies the correct execution of $F_{\text{priv}}$, without revealing its internal structure or parameters, while treating surrounding components with lighter-weight trust mechanisms. This decomposition allows teams to balance proof cost, model fidelity, and intellectual property protection.

This selective verification is embedded within a broader \emph{pragmatic graybox architecture}. {\projectName} does not enforce full transparency or full secrecy. Instead, it enables mixed execution environments in which some components are openly run, others are cryptographically secured, and the rest are entrusted to context-sensitive operational or economic safeguards.

\subsection{Fidelity and Model Degradation}
\label{sec:fidelity}

Before a model can participate in a ZKP, its floating-point computation must be \emph{circuit adapted}: weights and activations are quantized to fixed-point field elements, nonlinearities are replaced by low-degree surrogates, and the graph is re-expressed in terms of arithmetic-gate constraints. The resulting \emph{circuit-adapted model} differs from a conventional ``quantized'' model in that it typically undergoes a broader set of structural and numerical modifications driven by the requirements of finite-field computation. Each modification introduces a bounded distortion, but their cumulative effect can noticeably alter the model's output distribution---especially in deep or highly nonlinear networks.

We use the term \emph{fidelity} to refer to the proximity of a circuit-adapted model's outputs to those of its original floating-point counterpart. Fidelity is a measure of internal consistency, not predictive accuracy with respect to ground-truth labels. In this work, we quantify fidelity at the level of the pre-softmax \emph{logits}. While some proving systems may support softmax natively, our in-house research-stage prover JSTprove does not currently include softmax in its supported circuit components. For consistency across systems, we restrict our fidelity analysis to the pre-softmax \emph{logits}.

Given an input $x$ that produces logits
\[
\mathbf{z}^{\text{orig}}(x) = (y^{\text{orig}}_1,\ldots,y^{\text{orig}}_k), \quad \mathbf{z}^{\text{circ}}(x) = (y^{\text{circ}}_1,\ldots,y^{\text{circ}}_k),
\]
where the superscripts indicate the original and circuit-adapted models, we define the \emph{discrepancy} between the two as
\begin{equation}
\label{eq:def_discrepancy}
D_p(x) = \|\mathbf{z}^{\text{orig}}(x) - \mathbf{z}^{\text{circ}}(x)\|_p^p = \sum_{j = 1}^{k} \left|y^{\text{orig}}_j - y^{\text{circ}}_j\right|^p,
\end{equation}
where $p \in \{1, 2\}$ controls the sensitivity of the metric. Choosing $p = 1$ weights all coordinate-wise deviations equally, while $p = 2$ penalizes larger discrepancies more heavily. The latter is useful when fidelity loss is concentrated in a few coordinates, as it emphasizes large distortions more sharply. Normalizing by $1/k$ can be applied if fidelity comparisons across models with different output dimensions are needed.

While our fidelity analysis centers on logit-level discrepancies, we additionally assess the proximity of the full softmax output vectors. Although the softmax layer is not incorporated into the circuit itself, softmax-level comparisons still offer insight into how circuit adaptation and slicing affect the model's output distribution. In particular, they serve as a proxy for fidelity in applications where the final prediction depends on normalized probabilities.

Given two discrete probability distributions $P = (p_1,\ldots,p_k)$ and $Q = (q_1,\ldots,q_k)$ over $k$ classes, we consider two standard divergence measures. The \emph{total variation distance} (TVD) is defined by
\begin{equation}
\label{eq:TVD}
\operatorname{TVD}(P, Q) = \frac{1}{2} \sum_{i = 1}^{k} |p_i - q_i|,
\end{equation}
which measures the maximum amount of probability mass that must be shifted to transform one distribution into the other. It ranges from $0$ (identical distributions) to $1$ (disjoint support).

The \emph{Jensen--Shannon divergence} (JSD), a symmetric and smoothed version of the Kullback--Leibler divergence, is defined by
\begin{multline}
\label{eq:JSD}
\operatorname{JSD}(P, Q) =
\frac{1}{2} \sum_{i = 1}^{k} p_i \log_2 \left( \frac{2p_i}{p_i + q_i} \right) \\
+ \frac{1}{2} \sum_{i=1}^{k} q_i \log_2 \left( \frac{2q_i}{p_i + q_i} \right),
\end{multline}
where the base-$2$ logarithm ensures the divergence is measured in bits. JSD is always finite, bounded between $0$ and $1$ (if base-$2$ is used), and captures the similarity between distributions even when their supports differ.

\section{Threat and Trust Model}

{\projectName} is a framework for decentralized ML inference that aims to \emph{minimize trust} through strategic cryptographic verification. By enabling ZKPs of specific subcomputations, {\projectName} allows developers to reduce the trust surface and isolate critical components for formal validation. This supports flexible deployment strategies, where cost, risk, and verifiability can be balanced with fine-grained control.

\subsection*{Dual Trust Axes}

{\projectName} mediates between two distinct trust perspectives: the \emph{model provider} and the \emph{verifier}. The model provider seeks to protect proprietary models from theft or misuse. {\projectName} enables selective circuitization of the model, allowing sensitive components to be proven in zero-knowledge while leaving other parts open. This makes it possible to demonstrate correct execution of specific subcomputations without revealing weights, offering practical IP protection during R\&D phases, limited-access deployments, or commercial scenarios where full-model secrecy is either impractical or unnecessary. As with any deployed inference system, repeated interactions may reveal information about the underlying model, even if critical components are never directly exposed. {\projectName} does not attempt to eliminate this risk entirely. Rather, it mitigates leakage by isolating sensitive subcomputations, proving their correctness in zero-knowledge, and withholding internal parameters from the execution trace. While determined adversaries may still mount extraction attempts, the modular design and limited exposure surface raise the cost and complexity of such attacks.

The verifier, meanwhile, seeks assurance that the outputs they receive are the result of a faithful inference. When only part of the model is proven, this assurance becomes partial: each circuitized segment is cryptographically sound, but the correctness of the overall computation, including the coherence between inputs and outputs across unproven slices, relies on additional assumptions. These may include manual validation, redundant checks, or trust in the orchestration layer to preserve consistency. While this introduces some residual trust, {\projectName} allows that trust to be clearly scoped and, where possible, reduced.

\subsection*{Trust Boundaries and Strategic Verification}

Each circuitized slice defines a localized \emph{trust boundary}, within which the verifier can confidently check that a specific computation was performed correctly with respect to a fixed circuit and declared inputs and outputs. {\projectName} makes these boundaries explicit, supporting \emph{strategic verification} of critical components while permitting open execution of less sensitive parts.

When a slice is not circuitized, it may still be subject to scrutiny: the verifier can audit the computation directly, or rely on supporting mechanisms such as reproducibility, transparency, or a network of nodes whose behavior is constrained by incentives or the risk of detection. This flexibility allows {\projectName} to support a range of deployment models, from fully auditable pipelines to economically motivated orchestration.

\subsection*{Towards Composability}

Each slice in {\projectName} can be verified independently, providing localized assurance of correctness. While the inference as a whole is not yet cryptographically unified, the system is designed to support a practical middle ground: partial verification for high-leverage subcomputations, with consistency between slices maintained through auditability, incentives, or delegated trust in a network of compute nodes responsible for orchestrating intermediate inputs and outputs. This approach aligns with many real-world applications, where full end-to-end formal guarantees may be unnecessary or economically unjustifiable. At the same time, {\projectName} remains forward-compatible with more comprehensive cryptographic constructions, such as recursive proof composition or linking of intermediate states, for use cases that demand maximal assurance. 

\subsection*{Design Philosophy}

{\projectName} is built on the principle of \emph{strategic verifiability}: enabling strong guarantees where they matter most, while maintaining flexibility elsewhere. Rather than enforcing full cryptographic verification across the entire inference pipeline, {\projectName} focuses on verifying selected subcomputations: those that are particularly sensitive, proprietary, or security-critical. These verified segments can then be composed with unverified components using external mechanisms such as audit, redundancy, or replication. This enables a hybrid approach that balances formal assurances with practical feasibility.

In real-world deployments, full-model verification often remains out of reach due to computational constraints. Instead, developers can use {\projectName} to wrap critical portions of a model in ZKPs while treating other parts with lighter-weight trust models. For example, a cloud-based ML platform serving financial or medical applications might verify the model's final risk scoring or diagnostic output using ZKPs, while skipping earlier stages such as feature normalization or embedding lookup. These early stages can be recomputed or audited post hoc to ensure consistency.

Likewise, in an autonomous vehicle system, a full end-to-end cryptographic proof of inference may be infeasible in real time. However, selective verification of safety-critical modules, like decision-making logic for emergency braking or obstacle avoidance, can provide meaningful assurance, while less critical components (e.g., logging telemetry or aggregating non-urgent sensor data) are left unverified to conserve compute. This doesn't yield a formal guarantee over the entire pipeline, but it does create cryptographically strong checkpoints that can be linked by other trust mechanisms.

In distributed settings, this strategy becomes even more powerful. For example, in a decentralized supply chain, participants might generate ZKPs only for inferences involving high-value shipments, while relying on auditing and replication to handle less critical transactions. Here, {\projectName}'s modular design supports a spectrum of assurance strategies, from fully verified slices to loosely checked components, according to the risk profile of each task.

Selective verification allows organizations to balance performance, trust, and resource usage based on application-specific needs. The key tradeoff is ensuring the verification of high-risk, critical areas without overburdening the system with unnecessary computational load. 

\section{Experiments and Benchmarks}
\label{sec:experiments}

In this section, we implement and evaluate the performance of {\projectName} on a representative CNN model, the LeNet-5, a classic convolutional neural network introduced by LeCun et al.\ \cite{lecun1998gradient}, which still remains a widely used baseline in vision tasks due to its simplicity.

\subsection{Model Architecture}

Our implementation adapts the LeNet model for $3 \times 32 \times 32$ RGB inputs (e.g., CIFAR) and outputs a vector of 10 logits. The architecture consists of two convolutional layers with ReLU activations and max pooling, followed by three fully connected layers. The model is implemented in a modular format to enable slicing, with each major computational block structured as a standalone PyTorch module. This design facilitates flexible execution and targeted verification, aligning with {\projectName}'s architecture. Reference implementations and tutorials are available at~\cite{lenethf2025,lenetkaggle2025}.

\subsection{Slicing Strategy}

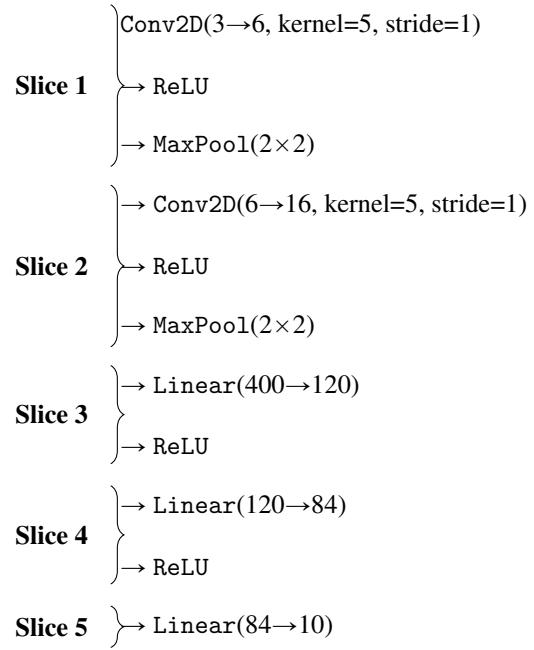
\begin{figure}[ht]
\centering
\[
\begin{tikzpicture}[baseline=(current bounding box.center)]

\tikzset{every node/.style={anchor=west}}

\node (a1) at (0,0)   {\texttt{Conv2D}(3$\rightarrow$6, kernel=5, stride=1)};
\node (a2) at (0,-0.8)   {$\rightarrow$ \texttt{ReLU}};
\node (a3) at (0,-1.6)   {$\rightarrow$ \texttt{MaxPool}(2$\times$2)};
\node (a4) at (0,-2.4)   {$\rightarrow$ \texttt{Conv2D}(6$\rightarrow$16, kernel=5, stride=1)};
\node (a5) at (0,-3.2)   {$\rightarrow$ \texttt{ReLU}};
\node (a6) at (0,-4.0)   {$\rightarrow$ \texttt{MaxPool}(2$\times$2)};
\node (a7) at (0,-4.8)   {$\rightarrow$ \texttt{Linear}(400$\rightarrow$120)};
\node (a8) at (0,-5.6)   {$\rightarrow$ \texttt{ReLU}};
\node (a9) at (0,-6.4)   {$\rightarrow$ \texttt{Linear}(120$\rightarrow$84)};
\node (a10) at (0,-7.2)  {$\rightarrow$ \texttt{ReLU}};
\node (a11) at (0,-8.0)  {$\rightarrow$ \texttt{Linear}(84$\rightarrow$10)};

\draw[decorate,decoration={brace,amplitude=5pt},xshift=-10pt]
  ($(a1.north west)$) -- ($(a3.south west)$) node [midway,xshift=-5pt,left] {\textbf{Slice 1}};

\draw[decorate,decoration={brace,amplitude=5pt},xshift=-10pt]
  ($(a4.north west)$) -- ($(a6.south west)$) node [midway,xshift=-5pt,left] {\textbf{Slice 2}};

\draw[decorate,decoration={brace,amplitude=5pt},xshift=-10pt]
  ($(a7.north west)$) -- ($(a8.south west)$) node [midway,xshift=-5pt,left] {\textbf{Slice 3}};

\draw[decorate,decoration={brace,amplitude=5pt},xshift=-10pt]
  ($(a9.north west)$) -- ($(a10.south west)$) node [midway,xshift=-5pt,left] {\textbf{Slice 4}};

\draw[decorate,decoration={brace,amplitude=5pt},xshift=-10pt]
  ($(a11.north west)$) -- ($(a11.south west)$) node [midway,xshift=-5pt,left] {\textbf{Slice 5}};

\end{tikzpicture}
\]
\caption{LeNet model architecture decomposed into five slices. Each slice corresponds to a modular block used for independent circuit generation.}
\label{fig:lenet-slices}
\end{figure}

In our implementation, the LeNet model is decomposed into five distinct slices, each corresponding to a a natural architectural block of computation: two convolutional blocks (each comprising a convolution, ReLU activation, and max pooling), followed by three fully connected blocks (two with ReLU activations), and a final output layer. Each slice is implemented as a standalone PyTorch module, enabling independent circuit generation and benchmarking along intuitive architectural boundaries. See Figure \ref{fig:lenet-slices}. 

\subsection{Proving Systems}

{\projectName} is designed as a prover-agnostic framework: it does not prescribe any particular proving system, but instead exposes interfaces that allow a wide range of proving systems to be plugged in. This modularity enables {\projectName} to support both established libraries and experimental protocols, and ensures that its benchmarking results reflect architectural properties of {\projectName} itself---not artifacts of a specific backend.

To demonstrate this flexibility, we benchmark {\projectName} using two distinct proving systems:
\begin{itemize}[noitemsep, topsep=0pt]
\item \textbf{EZKL} \cite{ezkl2025} --- a Halo2-based proving system with mature tooling, including a public CLI and support for ONNX model import;

\item \textbf{JSTprove} --- an internal, research-stage proving system built using Polyhedra Network's Expander Compiler Collection \cite{expandercc2025}. We design custom arithmetic circuits for each layer of the neural network using publicly available blueprints~\cite{zkmlblueprints2025}, and compile these into the Expander format to enable scalable, layer-wise verification via GKR and sumcheck protocols.
\end{itemize}

While these systems differ in maturity, architecture, and cryptographic foundations, we do not attempt a head-to-head comparison. Each entails distinct design tradeoffs, supported model classes, and threat models. Our goal is to validate that {\projectName}'s modular architecture functions robustly across qualitatively different proving stacks.

\subsection{Evaluation Inputs and Data Preparation}
\label{sec:data-generation}

To evaluate {\projectName} across proving systems and slicing configurations, we require a consistent set of inputs to feed into the benchmarked models. In this study, we use inputs drawn from the CIFAR dataset, consisting of natural images of size $3 \times 32 \times 32 = 3072$. A fixed batch of {\numDataPoints} images is selected uniformly at random from the available pool of 10{,}000 CIFAR samples. These inputs are then preprocessed and formatted for compatibility with the original model and its circuit-adapted variants.

The LeNet model used in our experiments follows a classic convolutional architecture and was lightly trained for demonstration purposes. While we do not report formal accuracy metrics, the model produces outputs that exhibit recognizable class preferences and variation across real-world inputs. This supports their use in fidelity experiments, where semantically meaningful logits are essential for comparing sliced and unsliced inference.

We hypothesize that circuit slicing improves fidelity because each slice's circuit handles a smaller, more localized computation. This reduces the need for aggressive quantization, bounded polynomial approximations, and other circuit adaptations that may degrade fidelity in monolithic circuits. Our empirical results on real-world data are consistent with this intuition and suggest that the benefits of slicing persist under more realistic deployment conditions. Importantly, slicing is performed \emph{before} circuit adaptation, so that each slice can be individually quantized and approximated as needed. This preserves local numerical behavior more faithfully than adapting a monolithic circuit and slicing afterward, which would defeat the purpose of slicing for fidelity gains.

\subsection{Memory Measurement Methodology}
\label{sec:memory-method}

To evaluate the memory requirements of different proving strategies, we record memory usage during witness generation, proof generation, and verification. For EZKL, this is measured externally using a background monitoring thread that tracks peak resident set size (RAM), estimated swap usage, and their sum across relevant subprocesses. These values are collected using platform-specific tools (\texttt{ps}, \texttt{vmmap}, and \texttt{psutil}) and reported for both unsliced and sliced models; in the sliced case, we record the peak across all slices. JSTprove, by contrast, reports a single fixed value that reflects its internal memory allocation rather than observed runtime consumption. This value is not measured externally and does not vary with input. Accordingly, we report it as a static figure without summary statistics. All measurements were conducted on a development workstation and should be interpreted as approximate empirical benchmarks rather than hardware-independent guarantees.

\subsection{Timing Measurement Methodology}
\label{sec:time-data}

Timing is measured using wall-clock duration for each phase of the proving process: witness generation, proof generation, and, where applicable, verification. In the sliced setting, we additionally record per-slice timings to capture how computation is distributed across the model. For EZKL, timing intervals begin immediately before the prover or verifier is invoked and end upon completion of the relevant process. These timings include all overhead, such as model loading, I/O, and preprocessing, to reflect realistic end-to-end performance. Measurements are collected externally through a unified orchestration layer. JSTprove timing data, by contrast, is reported using internal instrumentation built into the proving system. While there is some variation across inputs, the methodology differs from that used for EZKL. As such, timing comparisons between the two systems should be interpreted with care. As with memory, all timing results are hardware-dependent and should be treated as indicative rather than definitive.

\subsection{Fidelity Results}

We report fidelity statistics for each system using the distances $D_1$ and $D_2$, defined in Section~\ref{sec:fidelity} (see Equation~\eqref{eq:def_discrepancy}). These metrics quantify the discrepancy between the outputs of a circuit-adapted model and its original PyTorch counterpart. Results are shown in Tables~\ref{tab:ezkl_d1}--\ref{tab:JSTprove_d1_d2} for both unsliced and sliced variants, averaged over {\numDataPoints} CIFAR inputs. For EZKL, slicing yields a modest improvement in fidelity, with slightly lower mean values of $D_1$ and $D_2$. This aligns with the intuition that smaller circuit slices may require fewer approximations, producing outputs more consistent with the original model. In contrast, JSTprove shows no measurable fidelity difference between sliced and unsliced variants. The quantization step involves a user-chosen scaling factor that remains fixed across the model, but can be selected to balance range and precision. Overall, JSTprove exhibits slightly better fidelity to the original model than EZKL in our experiments, though this gap is small and context-dependent.

\FloatBarrier
\begin{table}[ht]
\centering
\begingroup
\setstretch{1.2}
\begin{tabular}{|l|l|l|}
\hline 
& 
$D_1$ Unsliced & 
$D_1$ Sliced \\
\hline
\texttt{mean} & 0.007615 & 0.007535 \\ \hline
 \texttt{std} & 0.003558 & 0.003500 \\ \hline
 \texttt{min} & 0.001836 & 0.001608 \\ \hline
 \texttt{max} & 0.018257 & 0.017590 \\ \hline
\end{tabular}
\endgroup
\caption{EZKL $D_1$ fidelity over {\numDataPoints} CIFAR inputs comparing circuit-adapted models to the original PyTorch model.}
\label{tab:ezkl_d1}
\end{table}
\FloatBarrier

\FloatBarrier
\begin{table}[ht]
\centering
\begingroup
\setstretch{1.2}
\begin{tabular}{|l|l|l|}
\hline 
& 
$D_2$ Unsliced & 
$D_2$ Sliced \\
\hline
\texttt{mean} & 1.011256e-05 & 9.925926e-06 \\ \hline
 \texttt{std} & 9.752333e-06 & 9.517990e-06 \\ \hline
 \texttt{min} & 4.896672e-07 & 4.571957e-07 \\ \hline
 \texttt{max} & 4.808309e-05 & 4.536161e-05 \\ \hline
\end{tabular}
\endgroup
\caption{EZKL $D_2$ fidelity over {\numDataPoints} CIFAR inputs comparing circuit-adapted models to the original PyTorch model.}
\label{tab:ezkl_d2}
\end{table}
\FloatBarrier

\FloatBarrier
\begin{table}[ht]
\centering
\begingroup
\setstretch{1.2}
\begin{tabular}{|l|l|l|}
\hline 
& 
$D_1$ (Un)sliced & 
$D_2$ (Un)sliced \\
\hline
\texttt{mean} & 0.001209 & 2.712345e-07 \\ \hline
 \texttt{std} & 0.000693 & 3.034757e-07 \\ \hline
 \texttt{min} & 0.000235 & 1.003568e-08 \\ \hline
 \texttt{max} & 0.003340 & 1.494044e-06 \\ \hline
\end{tabular}
\endgroup
\caption{JSTprove $D_1$, $D_2$ fidelity over {\numDataPoints} CIFAR inputs comparing circuit-adapted models to the original PyTorch model. Since slicing preserves logits exactly in this case, sliced and unsliced variants are identical.}
\label{tab:JSTprove_d1_d2}
\end{table}
\FloatBarrier

We additionally report divergence statistics between the softmax outputs of the circuit-adapted and original models, using total variation distance and Jensen--Shannon divergence as defined in Section~\ref{sec:fidelity}, Equations~\eqref{eq:TVD} and~\eqref{eq:JSD}. These metrics serve as a proxy for output-level fidelity in settings where normalized probability vectors are relevant. While the softmax layer itself is not circuitized (JSTprove does not yet support it), this comparison still provides a meaningful view into how circuit adaptation affects downstream outputs. If softmax were included in the circuit, such divergences would be the natural objects to measure. As Tables~\ref{tab:ezkl_tvd}--\ref{tab:JSTprove_tvd_js} show, the observed divergences remain negligible across all {\numDataPoints} CIFAR inputs tested, indicating that the numerical perturbations introduced by circuit adaptation and slicing do not meaningfully distort the model's confidence profile.

Finally, we verified that the predicted class, defined as the index of the maximum softmax entry, was invariant under circuit adaptation. Across all inputs, the sliced and unsliced variants of each proving system produced identical class predictions to the original PyTorch model. This further supports the conclusion that the loss of fidelity in our current setting is minimal and does not affect downstream decision-making. However, we emphasize that these results reflect a small, low-capacity model with minimal adaptation. In larger and more complex architectures, the cumulative effects of circuit adaptation may introduce more significant distortions. This is a direction that we plan to explore in future work.

\FloatBarrier
\begin{table}[ht]
\centering
\begingroup
\setstretch{1.2}
\begin{tabular}{|l|l|l|}
\hline 
& 
TVD Unsliced & 
TVD Sliced \\
\hline
\texttt{mean} & 0.000242 & 0.000238 \\ \hline
 \texttt{std} & 0.000211 & 0.000207 \\ \hline
 \texttt{min} & 0.000005 & 0.000005 \\ \hline
 \texttt{max} & 0.001167 & 0.001134 \\ \hline
\end{tabular}
\endgroup
\caption{EZKL total variation distance between softmax probability vectors of the circuit-adapted model and the original PyTorch model, across {\numDataPoints} CIFAR inputs.}
\label{tab:ezkl_tvd}
\end{table}
\FloatBarrier

\FloatBarrier
\begin{table}[ht]
\centering
\begingroup
\setstretch{1.2}
\begin{tabular}{|l|l|l|}
\hline 
& 
JS Unsliced & 
JS Sliced \\
\hline
\texttt{mean} & 1.063859e-07 & 1.026284e-07 \\ \hline
 \texttt{std} & 1.867072e-07 & 1.778216e-07 \\ \hline
 \texttt{min} & 8.044644e-10 & 8.533823e-10 \\ \hline
 \texttt{max} & 1.199443e-06 & 1.124447e-06 \\ \hline
\end{tabular}
\endgroup
\caption{EZKL Jensen-Shannon divergence between softmax probability vectors of the circuit-adapted model and the original PyTorch model, across {\numDataPoints} CIFAR inputs.}
\label{tab:ezkl_js}
\end{table}
\FloatBarrier

\FloatBarrier
\begin{table}[ht]
\centering
\begingroup
\setstretch{1.2}
\begin{tabular}{|l|l|l|}
\hline 
& 
TVD (Un)sliced & 
JS (Un)sliced \\
\hline
\texttt{mean} & 0.000026 & 1.070359e-09 \\ \hline
 \texttt{std} & 0.000012 & 7.375106e-10 \\ \hline
 \texttt{min} & 0.000007 & 1.463187e-10 \\ \hline
 \texttt{max} & 0.000056 & 2.946555e-09 \\ \hline
\end{tabular}
\endgroup
\caption{JSTprove total variation distance and Jensen-Shannon divergence between softmax probability vectors of the circuit-adapted model and the original PyTorch model. Since slicing preserves logits exactly in this case, sliced and unsliced variants are identical.}
\label{tab:JSTprove_tvd_js}
\end{table}
\FloatBarrier

\subsection{Memory Usage Results}
\label{sec:memory-data}

While our evaluation includes both sliced and unsliced variants, we caution against direct comparisons of their memory footprints. The unsliced case corresponds to a full proof of inference, in which a single monolithic circuit spans the entire model. In contrast, the sliced configuration consists of multiple independent proofs, each covering a portion of the computation. These sliced circuits are not combined into a single global proof and should not be viewed as a substitute for end-to-end verifiability in a purely formal cryptographic sense. For EZKL, peak memory usage was measured dynamically and varies across inputs. The results indicate that slicing leads to a substantial reduction in peak memory, particularly during proof generation, compared to the monolithic case. In scenarios where full inference verification is unnecessary, slicing offers a pragmatic tradeoff: reduced memory requirements in exchange for architectural complexity and reliance on external mechanisms to ensure consistency across slices. See Table~\ref{tab:ezkl_memory} for details.

\FloatBarrier
\begin{table}[ht]
\centering
\begingroup
\setstretch{1.2}
\resizebox{\columnwidth}{!}{
\begin{tabular}{|l|l|l|l|l|}
\hline  
\texttt{Cfg/Stage} & \texttt{Stat} & \texttt{RAM} & \texttt{Swap} & \texttt{Sum} \\
\hline\hline

\multirow{4}{*}{\shortstack[l]{Full Inference \\ Witness}} 
    & \texttt{mean} & 1,048.574 & -- & 1,048.574 \\ \cline{2-5}
    & \texttt{std}  & 28.522 & -- & 28.522 \\ \cline{2-5}
    & \texttt{min}  & 1,036.359 & -- & 1,036.359 \\ \cline{2-5}
    & \texttt{max}  & 1,215.344 & -- & 1,215.344 \\ \hline

\multirow{4}{*}{\shortstack[l]{Per-slice\\ Witness}} 
    & \texttt{mean} & 51.579 & -- & 51.579 \\ \cline{2-5}
    & \texttt{std}  & 7.461 & -- & 7.461 \\ \cline{2-5}
    & \texttt{min}  & 27.922 & -- & 27.922 \\ \cline{2-5}
    & \texttt{max}  & 56.234 & -- & 56.234 \\ \hline\hline

\multirow{4}{*}{\shortstack[l]{Full Inference \\ Proof}} 
    & \texttt{mean} & 4,854.582 & 27,408.091 & 32,262.673 \\ \cline{2-5}
    & \texttt{std}  & 507.130 & 1,452.543 & 1,327.700 \\ \cline{2-5}
    & \texttt{min}  & 3,613.781 & 24,166.399 & 29,809.806 \\ \cline{2-5}
    & \texttt{max}  & 5,958.016 & 30,924.800 & 35,402.581 \\ \hline

\multirow{4}{*}{\shortstack[l]{Per-slice \\ Proof}} 
    & \texttt{mean} & 7,636.568 & 12,469.394 & 20,105.962 \\ \cline{2-5}
    & \texttt{std}  & 998.847 & 726.000 & 1,076.843 \\ \cline{2-5}
    & \texttt{min}  & 5,422.344 & 11,161.600 & 17,441.024 \\ \cline{2-5}
    & \texttt{max}  & 8,832.859 & 14,540.800 & 22,268.641 \\ \hline\hline

\multirow{4}{*}{\shortstack[l]{Full Inference \\ Verification}} 
    & \texttt{mean} & 973.510 & -- & 973.510 \\ \cline{2-5}
    & \texttt{std}  & 342.043 & -- & 342.043 \\ \cline{2-5}
    & \texttt{min}  & 323.016 & -- & 323.016 \\ \cline{2-5}
    & \texttt{max}  & 1,375.516 & -- & 1,375.516 \\ \hline

\multirow{4}{*}{\shortstack[l]{Per-slice \\ Verification}} 
    & \texttt{mean} & 536.216 & -- & 536.216 \\ \cline{2-5}
    & \texttt{std}  & 141.856 & -- & 141.856 \\ \cline{2-5}
    & \texttt{min}  & 5.938 & -- & 5.938 \\ \cline{2-5}
    & \texttt{max}  & 660.719 & -- & 660.719 \\ \hline
\end{tabular}
}
\endgroup
\caption{EZKL peak memory usage (in megabytes) measured externally across {\numDataPoints} CIFAR inputs.}
\label{tab:ezkl_memory}
\end{table}

For JSTprove, memory usage is reported as a fixed internal allocation value, independent of input. While this value provides insight into the prover's internal design, it does not reflect observed memory consumption and should not be directly compared to the dynamic measurements collected for EZKL. For EZKL, slicing provides substantial memory benefits during witness generation, with per-slice usage approximately 20 times lower than in the monolithic case. For proof generation, total memory usage drops by approximately 38\% in the sliced configuration. Verification memory also improves under slicing, with mean usage falling by roughly 45\%. Values are shown for RAM, swap, and their sum, across different stages (witness, proof, verification) and configurations (full inference, per-slice). RAM and swap peaks may occur at different times; the ``Sum'' column provides a loose upper bound assuming worst-case simultaneous peak allocation. Since our evaluation was conducted on a compact convolutional model, we expect the relative benefits of slicing, particularly in terms of swap reduction, to become more pronounced in some cases as model size increases. However, peak memory usage per slice may still become a bottleneck for deep or high-resolution architectures, underscoring the importance of continued improvements in prover efficiency.

\FloatBarrier
\begin{figure}[ht]
\centering
\includegraphics[width=0.8\columnwidth]{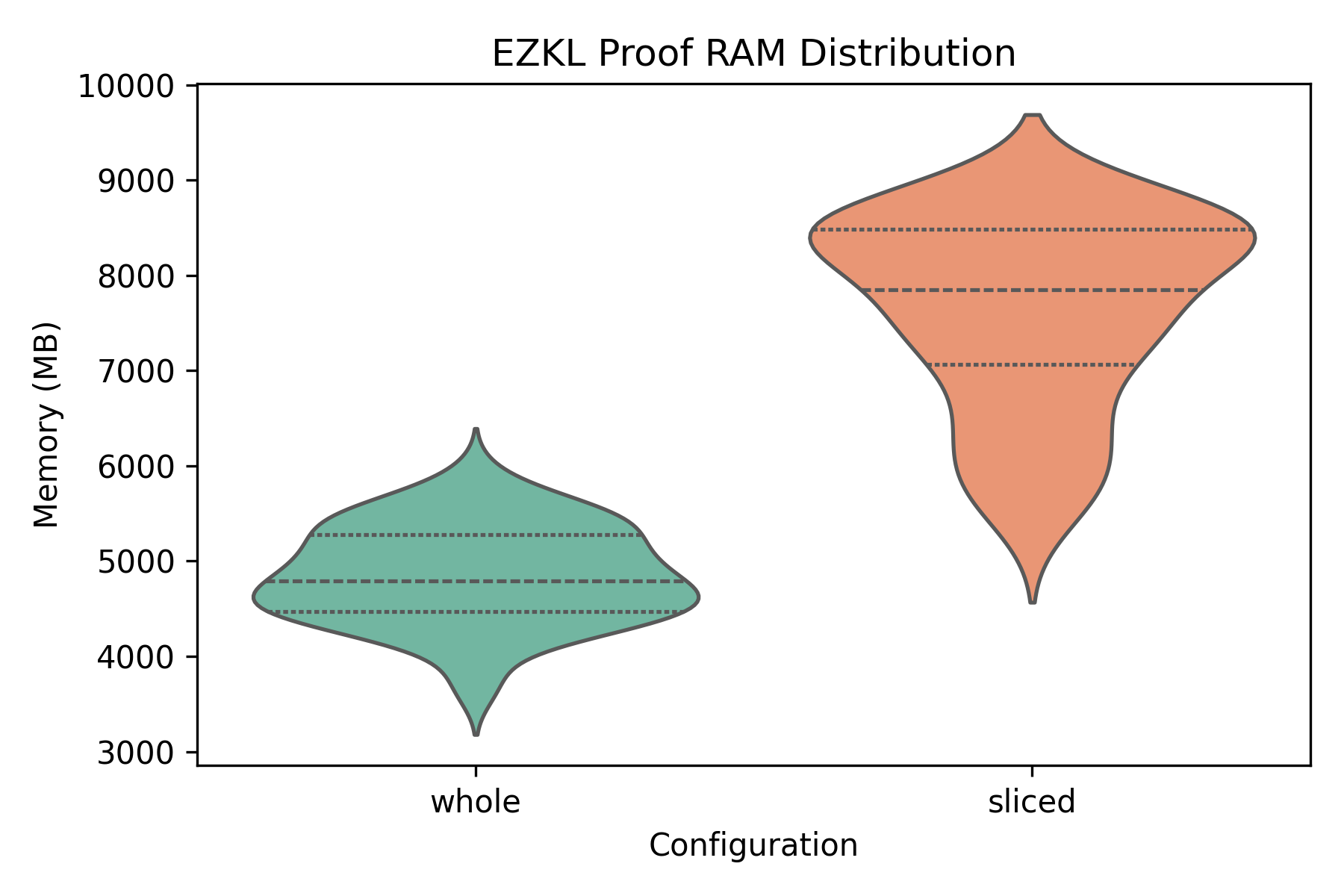}
\caption{EZKL proof-stage peak RAM usage across \texttt{whole}/full inference and \texttt{sliced} configurations.}
\label{fig:ezkl_proof_ram_violin}
\end{figure}
\FloatBarrier

\FloatBarrier
\begin{figure}[ht]
\centering
\includegraphics[width=0.8\columnwidth]{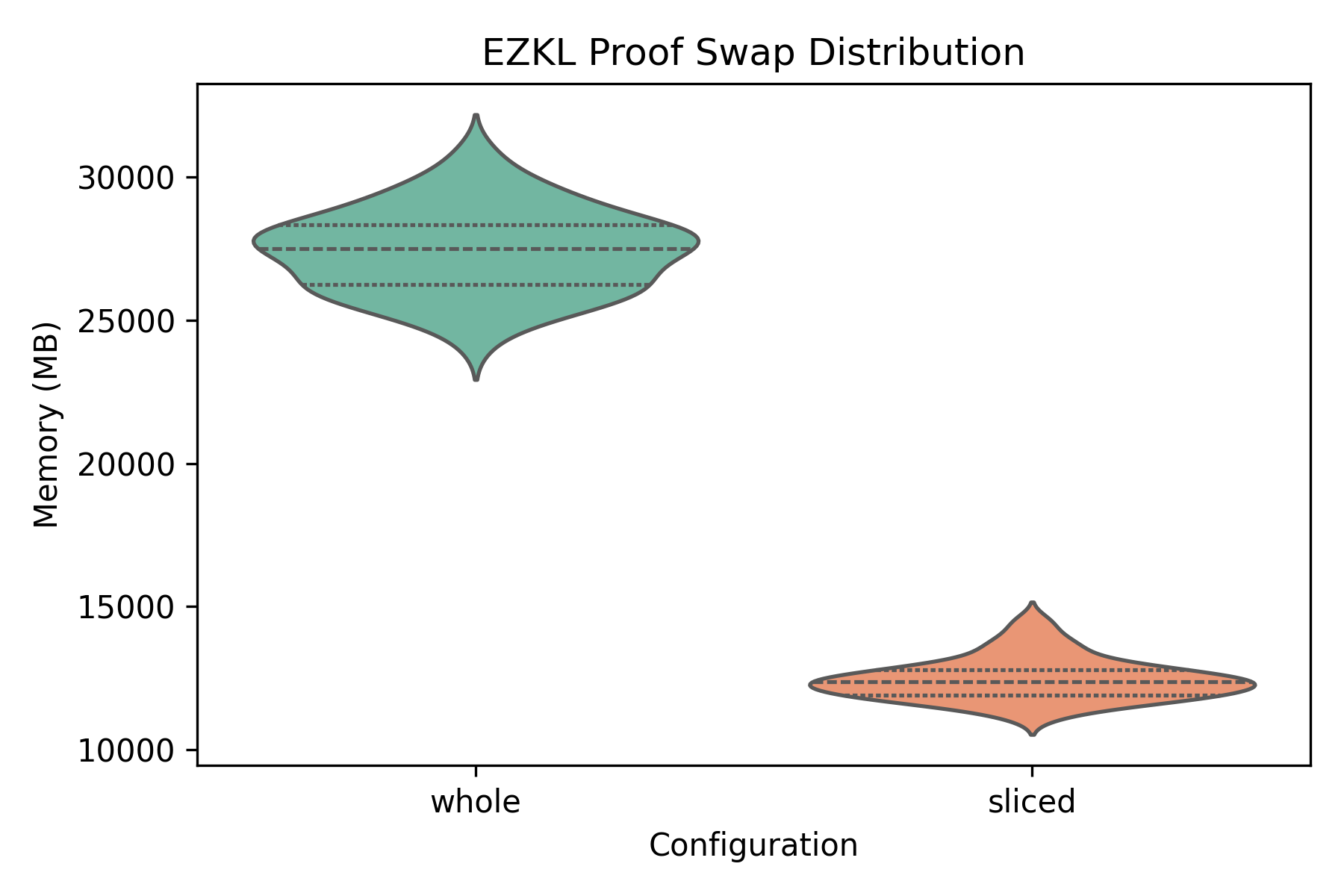}
\caption{EZKL proof-stage peak swap usage across \texttt{whole}/full inference and \texttt{sliced} configurations.}
\label{fig:ezkl_proof_swap_violin}
\end{figure}
\FloatBarrier

\FloatBarrier
\begin{figure}[ht]
\centering
\includegraphics[width=0.8\columnwidth]{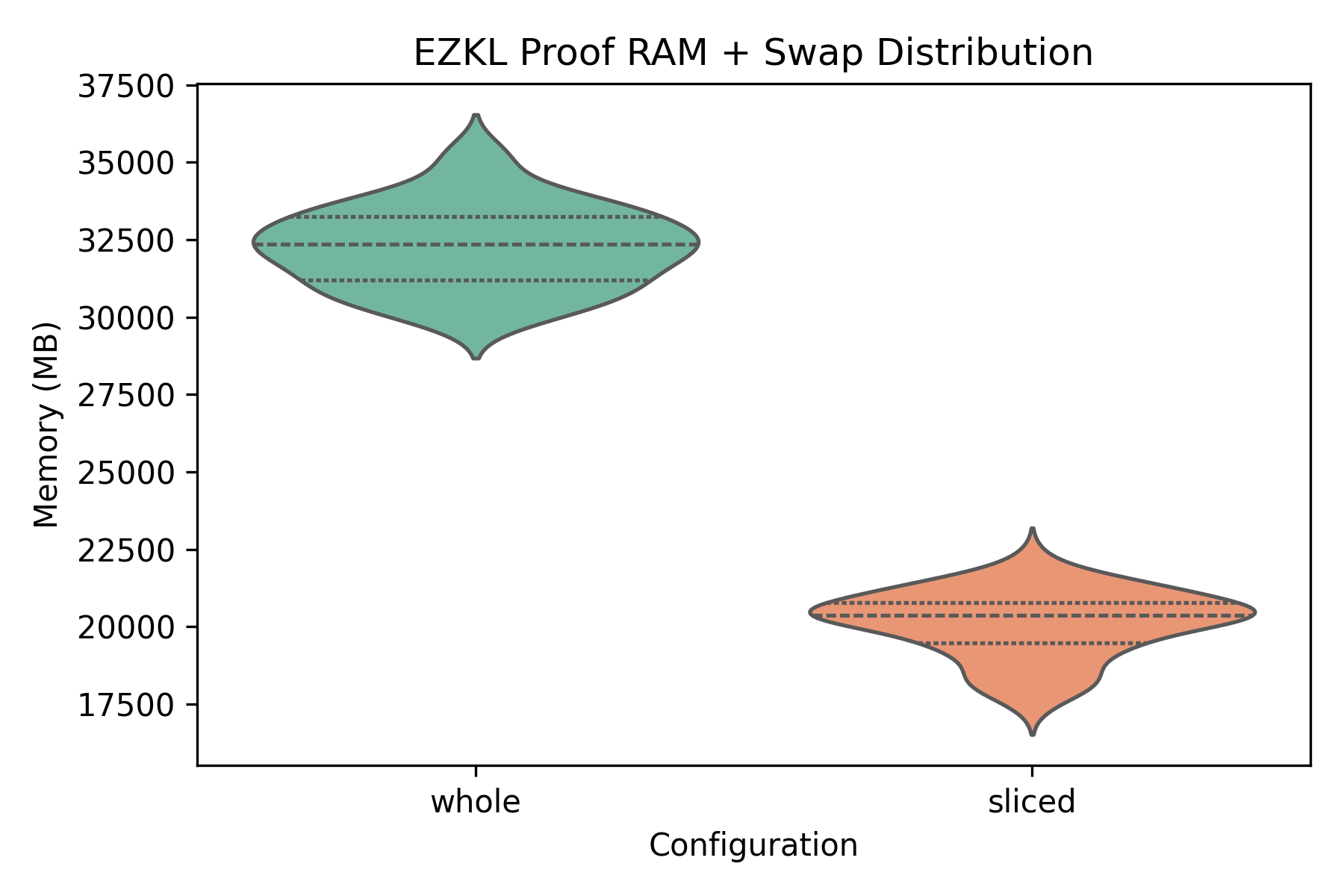}
\caption{EZKL proof-stage total memory usage across \texttt{whole}/full inference and \texttt{sliced} configurations.}
\label{fig:ezkl_proof_total_violin}
\end{figure}
\FloatBarrier

For JSTprove, the prover reports a fixed memory allocation value for each stage of execution, rather than tracking dynamic memory usage. Under slicing, the allocated memory is approximately halved across all stages: witness generation, proof generation, and verification (see Table \ref{tab:JSTprove_memory} for details). This likely reflects reduced buffer sizes and intermediate data structures when handling smaller circuits per slice. However, because these values represent internal allocation rather than observed consumption, and remain constant across inputs, they should be interpreted as indicative of internal prover design rather than empirical performance. These values are internally reported by the prover and reflect allocated memory, not observed peak usage. They should not be compared directly to the externally measured EZKL results.

We anticipate that the relative memory savings observed here may scale favorably with larger models, as the overhead associated with monolithic circuit construction and execution grows with network depth and width.

\FloatBarrier
\begin{table}[ht]
\centering
\begingroup
\setstretch{1.2}
\begin{tabular}{|l|l|c|}
\hline  
\texttt{Cfg/Stage} & \texttt{Allocated} \\
\hline\hline
Full Inference Witness     & 160.310 \\ \hline
Per-slice Witness       & 80.510  \\ \hline
Full Inference Proof       & 1,084.960 \\ \hline
Per-slice Proof         & 544.280    \\ \hline
Full Inference Verification & 1,082.620  \\ \hline
Per-slice Verification   & 543.110   \\ \hline
\end{tabular}
\endgroup
\caption{JSTprove memory allocation (in megabytes) for each stage (witness, proof, verification) under both unsliced and sliced configurations.}
\label{tab:JSTprove_memory}
\end{table}
\FloatBarrier

\subsection{Timing Results}
\label{subsec:time-results}

As with memory, we avoid direct comparisons between the total runtime of sliced and unsliced configurations. The unsliced case reflects the time required to prove a complete model inference using a single monolithic circuit, whereas the sliced case partitions the computation into five independently proved subcircuits. Since each slice is verified in isolation, the resulting execution does not constitute a cryptographic proof of full inference, but rather a collection of localized claims.

Timing results for EZKL were measured externally using wall-clock timing from the orchestration layer, while JSTprove reports timing using internal instrumentation. Although these methods are not directly comparable, both systems exhibit reduced witness and proof generation times when inference is decomposed into smaller subcomputations. However, these improvements are only meaningful in contexts where verifying individual slices is sufficient, and full end-to-end consistency is not required. In such settings, slicing can improve prover throughput and responsiveness, provided the system can tolerate weaker formal guarantees and additional complexity in orchestration.

Slicing substantially reduces runtime for witness and proof generation in EZKL, but only under the assumption that isolated slice-level verification is sufficient. Witness generation time drops by roughly 77\% in the per-slice setting, while proof generation time decreases by approximately 66\%. These gains reflect the smaller circuit size and lower computational burden associated with proving just a portion of the model. However, such reductions are only meaningful when full end-to-end cryptographic guarantees are unnecessary. 

Verification time shows a more modest improvement, falling by about 38\%, but this figure does not include any logic for enforcing consistency across slices. In scenarios where intermediate values must be stitched together or verified jointly, additional overhead would be required. As such, the timing benefits of slicing should be interpreted as localized optimizations rather than systemic improvements in full-model verifiability.

That said, we anticipate that the relative gains from slicing may become more pronounced in larger or deeper networks, where monolithic circuit construction and proving cost grow disproportionately with model size, while the per-slice footprint may remain bounded if slices are kept shallow. For sliced configurations, ``Total'' time (see Table \ref{tab:ezkl_time} for details) reflects the sum of individual slice executions including orchestration overhead. Verification times do not include consistency checks or output chaining across slices.

\FloatBarrier
\begin{table}[ht]
\centering
\begingroup
\setstretch{1.2}
\resizebox{\columnwidth}{!}{
\begin{tabular}{|l|l|l|l|l|l|}
\hline  
\texttt{Cfg/Stage} & \texttt{Slice} & \texttt{mean} & \texttt{std} & \texttt{min} & \texttt{max} \\
\hline\hline

\makecell[l]{Full Inference \\ Witness} 
  & -- & 1.584 & 0.211 & 1.386 & 2.194 \\ \hline

\multirow{8}{*}{\shortstack{Per-slice \\ Witness}} 
    & \texttt{Total}   & 0.364 & 0.154 & 0.284 & 0.996 \\ \cline{2-6}
    & \texttt{Slice 1} & 0.215 & 0.146 & 0.147 & 0.798 \\ \cline{2-6}
    & \texttt{Slice 2} & 0.080 & 0.007 & 0.070 & 0.118 \\ \cline{2-6}
    & \texttt{Slice 3} & 0.034 & 0.006 & 0.029 & 0.074 \\ \cline{2-6}
    & \texttt{Slice 4} & 0.021 & 0.003 & 0.017 & 0.031 \\ \cline{2-6}
    & \texttt{Slice 5} & 0.012 & 0.003 & 0.010 & 0.025 \\ \hline\hline

\makecell[l]{Full Inference \\ Proof} 
  & -- & 478.683 & 41.558 & 425.110 & 643.144 \\ \hline

\multirow{8}{*}{\shortstack[l]{Per-slice \\ Proof}} 
    & \texttt{Total}   & 163.182 & 11.987 & 151.067 & 215.335 \\ \cline{2-6}
    & \texttt{Slice 1} & 111.474 & 9.830  & 102.706 & 154.871 \\ \cline{2-6}
    & \texttt{Slice 2} & 43.620  & 2.258  & 40.527  & 53.921  \\ \cline{2-6}
    & \texttt{Slice 3} & 2.852   & 0.334  & 1.066   & 3.897   \\ \cline{2-6}
    & \texttt{Slice 4} & 2.651   & 0.371  & 0.960   & 3.962   \\ \cline{2-6}
    & \texttt{Slice 5} & 2.584   & 0.286  & 0.971   & 3.373   \\ \hline\hline

\makecell[l]{Full Inference \\ Verification} 
  & -- & 0.988 & 0.239 & 0.814 & 1.870 \\ \hline

\multirow{8}{*}{\shortstack[l]{Per-slice \\ Verification}} 
    & \texttt{Total}   & 0.605 & 0.206 & 0.371 & 0.976 \\ \cline{2-6}
    & \texttt{Slice 1} & 0.436 & 0.202 & 0.209 & 0.779 \\ \cline{2-6}
    & \texttt{Slice 2} & 0.113 & 0.008 & 0.101 & 0.137 \\ \cline{2-6}
    & \texttt{Slice 3} & 0.020 & 0.002 & 0.018 & 0.025 \\ \cline{2-6}
    & \texttt{Slice 4} & 0.018 & 0.001 & 0.016 & 0.021 \\ \cline{2-6}
    & \texttt{Slice 5} & 0.017 & 0.001 & 0.016 & 0.022 \\ \hline
\end{tabular}
}
\endgroup
\caption{
EZKL runtime (in seconds) across {\numDataPoints} CIFAR inputs for each stage (witness, proof, verification) and configuration (full inference vs.\ per-slice).}
\label{tab:ezkl_time}
\end{table}
\FloatBarrier

JSTprove exhibits a different performance profile than EZKL under slicing. Most notably, witness generation time increases significantly in the sliced configuration, rising from 0.27s to 1.52s. This suggests that the overhead introduced by orchestrating multiple circuits, or the internal structure of our circuit designs, may dominate runtime in smaller models. Proof generation time decreases modestly (by about 9\%), implying that slicing offers limited performance gains in this setting. This may reflect the scalability of Expander's underlying protocols, on which JSTprove is built, which are designed to perform efficiently even for moderately large circuits. Verification time closely mirrors proof time in both configurations, a behavior that appears consistent with internal characteristics of the Expander framework, as corroborated by prior experience. These results highlight the architectural cost of slicing when circuit-level efficiencies are already well optimized. It is possible that more substantial gains would emerge in larger models, where the overhead of proving a monolithic computation becomes more pronounced relative to smaller, modular slices.

\FloatBarrier
\begin{figure}[ht]
\centering
\includegraphics[width=0.8\columnwidth]{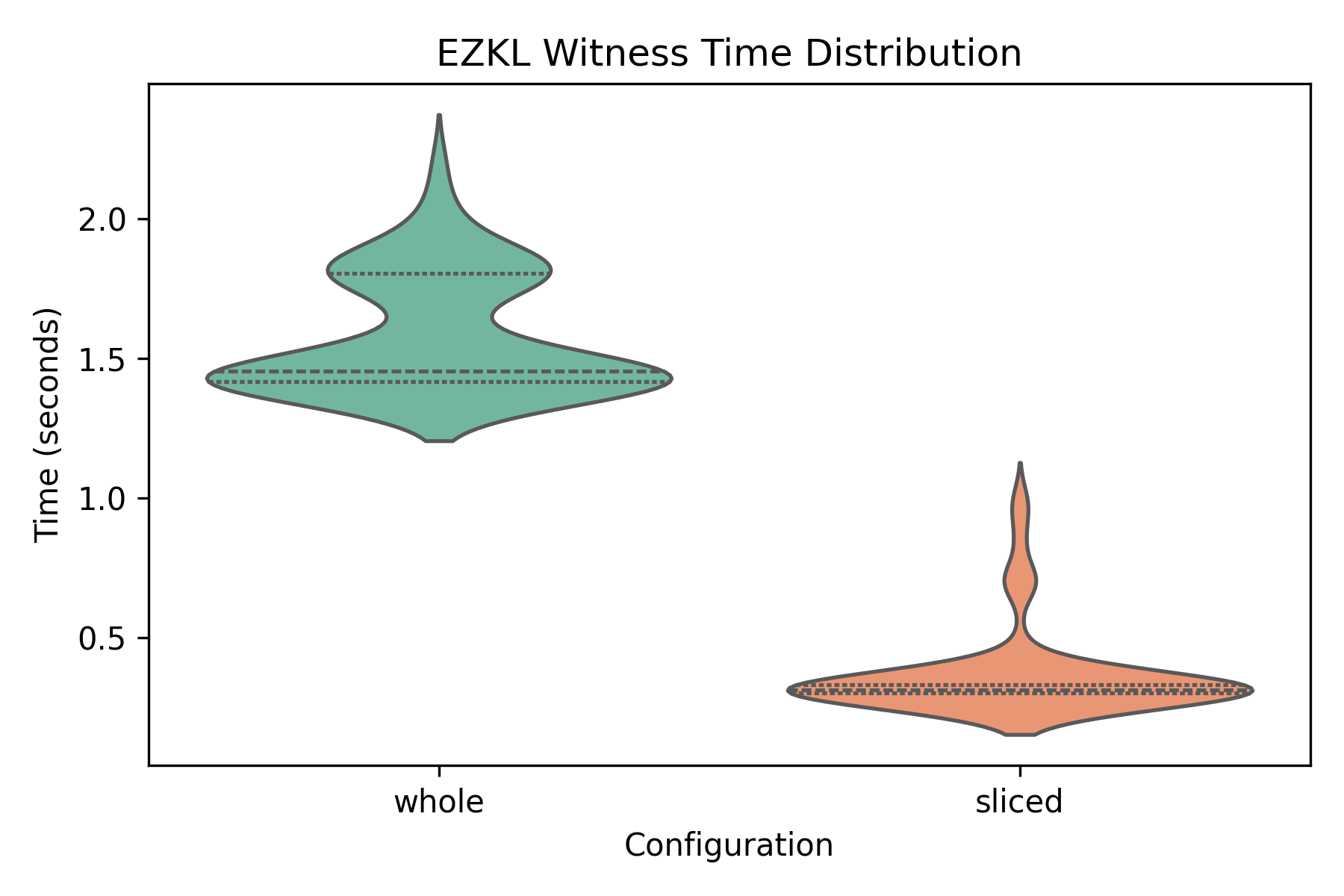}
\caption{EZKL witness generation time across \texttt{whole} (full inference) and \texttt{sliced} configurations.}
\label{fig:ezkl_witness_time_violin}
\end{figure}
\FloatBarrier

\FloatBarrier
\begin{figure}[ht]
\centering
\includegraphics[width=0.8\columnwidth]{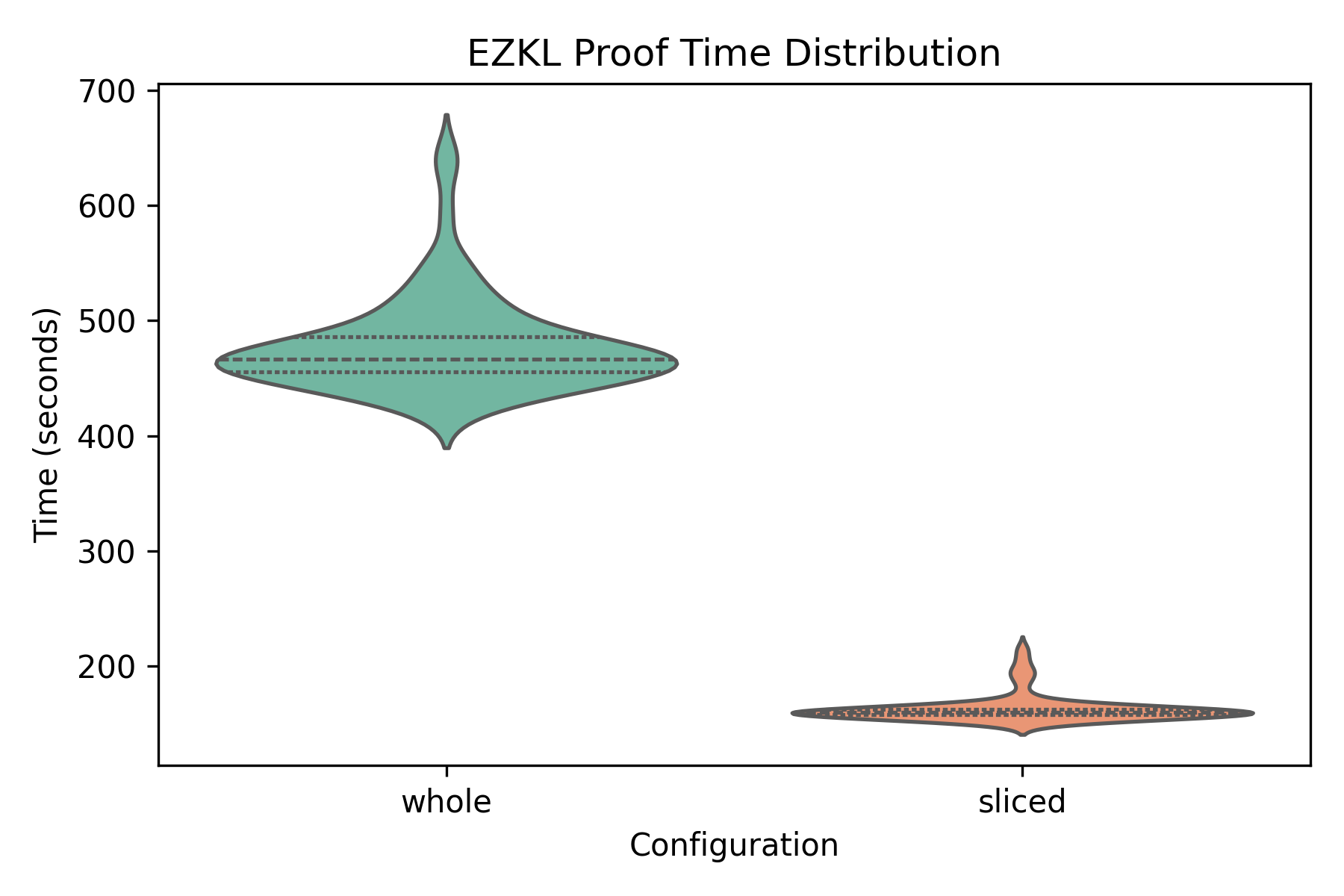}
\caption{EZKL proof generation time across \texttt{whole} (full inference) and \texttt{sliced} configurations.}
\label{fig:ezkl_proof_time_violin}
\end{figure}
\FloatBarrier

\FloatBarrier
\begin{figure}[ht]
\centering
\includegraphics[width=0.8\columnwidth]{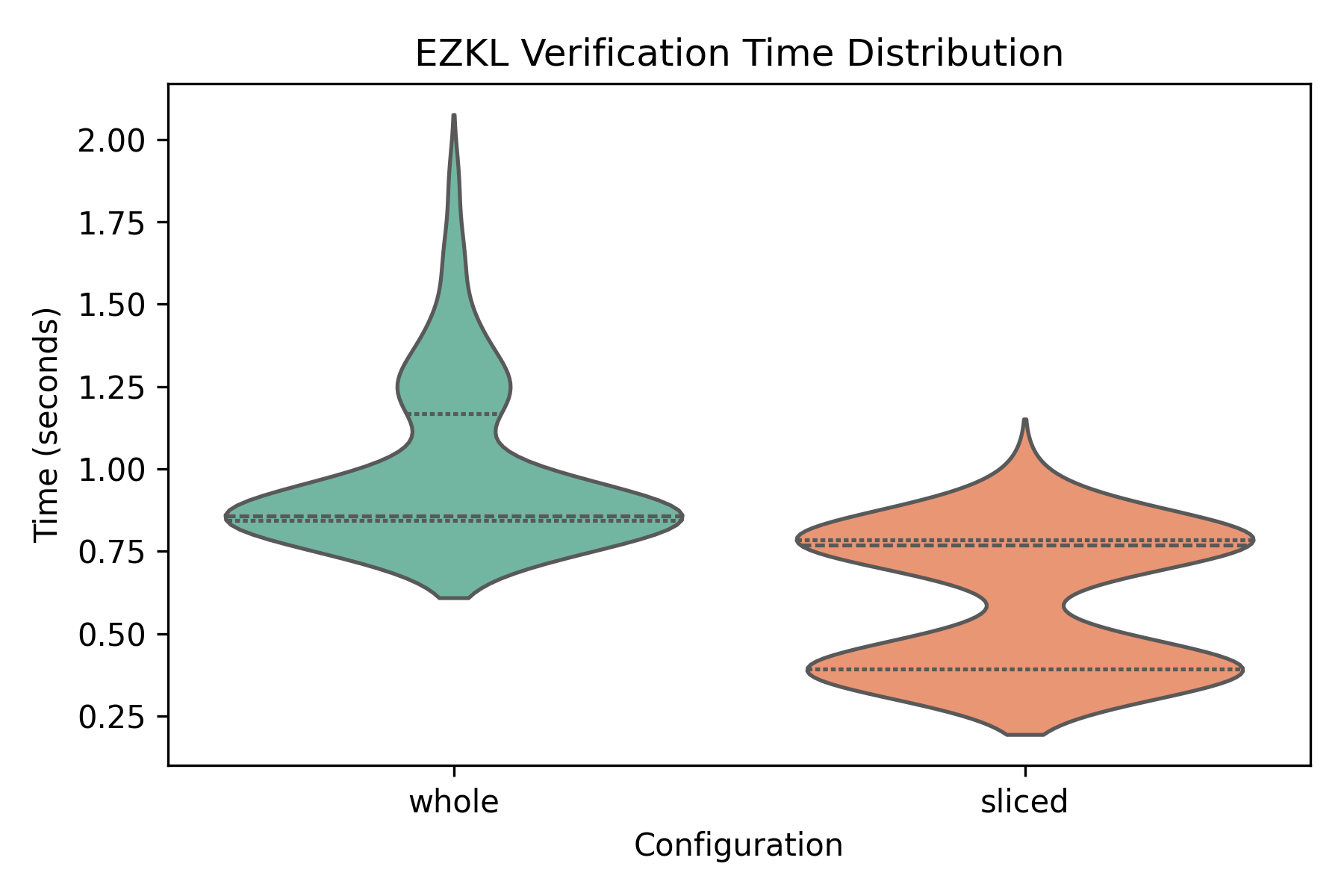}
\caption{EZKL verification time across \texttt{whole} (full inference) and \texttt{sliced} configurations. Note that verification times do not include consistency checks across slices.}
\label{fig:ezkl_verification_time_violin}
\end{figure}
\FloatBarrier

For sliced configurations, ``Total'' time (see Table \ref{tab:JSTprove_time} for details) includes end-to-end orchestration and prover runtime overhead, and may differ slightly from the sum of per-slice times due to measurement granularity. Verification times exclude input/output consistency checks across slices. We note that witness generation time for Slice 5 is reported as zero in JSTprove. This is likely a measurement artifact, possibly due to instrumentation behavior, implicit witness construction, or the extremely small size of the circuit involved. We did not investigate this further, as the impact on total runtime is negligible in either case.

\FloatBarrier
\begin{table}[ht]
\centering
\begingroup
\setstretch{1.2}
\resizebox{\columnwidth}{!}{
\begin{tabular}{|l|l|l|l|l|l|}
\hline  
\texttt{Cfg/Stage} & \texttt{Slice} & \texttt{mean} & \texttt{std} & \texttt{min} & \texttt{max} \\
\hline\hline

\makecell[l]{Full Inference \\ Witness} 
  & -- & 0.269 & 0.014 & 0.256 & 0.313 \\ \hline

\multirow{6}{*}{\shortstack[l]{Per-slice \\ Witness}} 
    & \texttt{Total}   & 1.523 & 0.057 & 1.451 & 1.727 \\ \cline{2-6}
    & \texttt{Slice 1} & 0.182 & 0.010 & 0.174 & 0.231 \\ \cline{2-6}
    & \texttt{Slice 2} & 0.736 & 0.043 & 0.670 & 0.860 \\ \cline{2-6}
    & \texttt{Slice 3} & 0.406 & 0.029 & 0.400 & 0.600 \\ \cline{2-6}
    & \texttt{Slice 4} & 0.200 & 0.000 & 0.200 & 0.200 \\ \cline{2-6}
    & \texttt{Slice 5} & 0.000 & 0.000 & 0.000 & 0.000 \\ \hline\hline

\makecell[l]{Full Inference \\ Proof} 
  & -- & 9.786 & 0.432 & 9.498 & 11.707 \\ \hline

\multirow{6}{*}{\shortstack[l]{Per-slice \\ Proof}} 
    & \texttt{Total}   & 8.872 & 0.342 & 8.672 & 10.446 \\ \cline{2-6}
    & \texttt{Slice 1} & 5.645 & 0.201 & 5.539 & 6.507 \\ \cline{2-6}
    & \texttt{Slice 2} & 2.479 & 0.194 & 2.409 & 3.720 \\ \cline{2-6}
    & \texttt{Slice 3} & 0.137 & 0.009 & 0.127 & 0.195 \\ \cline{2-6}
    & \texttt{Slice 4} & 0.125 & 0.008 & 0.119 & 0.163 \\ \cline{2-6}
    & \texttt{Slice 5} & 0.485 & 0.026 & 0.440 & 0.610 \\ \hline\hline

\makecell[l]{Full Inference \\ Verification} 
  & -- & 9.608 & 0.494 & 9.361 & 12.148 \\ \hline

\multirow{6}{*}{\shortstack[l]{Per Slice \\ Verification}} 
    & \texttt{Total}   & 8.694 & 0.321 & 8.540 & 10.707 \\ \cline{2-6}
    & \texttt{Slice 1} & 5.540 & 0.175 & 5.457 & 6.426 \\ \cline{2-6}
    & \texttt{Slice 2} & 2.424 & 0.139 & 2.377 & 3.427 \\ \cline{2-6}
    & \texttt{Slice 3} & 0.135 & 0.008 & 0.124 & 0.181 \\ \cline{2-6}
    & \texttt{Slice 4} & 0.123 & 0.007 & 0.118 & 0.162 \\ \cline{2-6}
    & \texttt{Slice 5} & 0.472 & 0.017 & 0.440 & 0.530 \\ \hline
\end{tabular}
}
\endgroup
\caption{
JSTprove runtime (in seconds) across {\numDataPoints} CIFAR inputs for each stage (witness, proof, verification) and configuration (full inference vs.\ per-slice).}
\label{tab:JSTprove_time} 
\end{table}
\FloatBarrier

\section{Limitations and Future Work}

While {\projectName} demonstrates that targeted verification is viable in real-world zkML pipelines, several limitations remain and suggest directions for future development. Our evaluation focuses on a small convolutional network; further work is needed to characterize fidelity, memory, and runtime behavior on larger, more complex models. Although slices can collectively span an entire inference, {\projectName} does not yet support compositional proof linking, and integrating recursive ZKPs remains a challenging problem. As with all systems that aim to preserve proprietary logic, long-term inference leakage and model confidentiality must be considered, especially in repeated or interactive settings.

At present, slicing strategies are specified manually. Automating this process, while balancing resource constraints, model semantics, and verifiability, may improve usability and performance. The guarantees provided by {\projectName} are localized to individual slices; maintaining consistency across the full inference relies on external orchestration, auditability, or incentive structures. While this is often sufficient in practice, a more formal understanding of deployment semantics would help clarify system-level guarantees. Looking ahead, {\projectName}'s modular architecture may serve as a foundation for full-model verification as recursive composition frameworks become more mature and efficient.

In parallel, we plan to apply a unified benchmarking methodology to JSTprove and other proving systems, to enable more consistent and comparable measurement of memory and timing performance across backends.

\section{Conclusion}

Verifiable inference remains a central goal for secure and trustworthy ML, particularly in decentralized and adversarial environments. Yet full-model circuitization remains prohibitively expensive for most practical deployments. {\projectName} offers a pragmatic alternative: a modular framework for selectively verifying high-value subcomputations through independently provable slices.

We have presented the design of {\projectName}, outlined its trust model and architectural constraints, and evaluated its performance on a small convolutional network using two distinct proving systems. As expected, slicing leads to substantial reductions in proof-generation time and memory requirements, without compromising fidelity. In fact, we observe marginal improvements in fidelity under slicing, an effect we expect to become more pronounced as model size and circuit complexity increase. These trends, along with scalability to larger architectures, remain an area for future investigation.

While slicing enables targeted verification, the scalability of this approach is fundamentally limited by the resource demands of individual layers. As models grow in size and complexity, even isolated segments may exceed available memory or compute budgets. This suggests that practical viability will depend not only on slicing strategies but also on improvements to circuit efficiency and proving backends.

{\projectName}'s modular design is compatible with existing zkML stacks and can be deployed today in settings where full inference verification is impractical. At the same time, it remains forward-compatible with emerging proof composition frameworks, offering a realistic foundation for verifiable ML pipelines both now and as cryptographic infrastructure matures.

\section{Acknowledgments}

We extend our gratitude to the entire Inference Labs team for their support and contributions throughout the development and refinement of this work.

\IEEEtriggeratref{9} 
\printbibliography[title=References,heading=bibintoc]

@article{ghodsi2017safetynets,
  title={Safetynets: Verifiable execution of deep neural networks on an untrusted cloud},
  author={Ghodsi, Zahra and Gu, Tianyu and Garg, Siddharth},
  journal={Advances in Neural Information Processing Systems},
  volume={30},
  year={2017}
}

@article{peng2025survey,
  title={A survey of zero-knowledge proof based verifiable machine learning},
  author={Peng, Zhizhi and Wang, Taotao and Zhao, Chonghe and Liao, Guofu and Lin, Zibin and Liu, Yifeng and Cao, Bin and Shi, Long and Yang, Qing and Zhang, Shengli},
  journal={arXiv preprint arXiv:2502.18535},
  year={2025}
}

@article{south2024verifiable,
  title={Verifiable evaluations of machine learning models using zkSNARKs},
  author={South, Tobin and Camuto, Alexander and Jain, Shrey and Nguyen, Shayla and Mahari, Robert and Paquin, Christian and Morton, Jason and Pentland, Alex'Sandy'},
  journal={arXiv preprint arXiv:2402.02675},
  year={2024}
}

@article{sheybani2025zero,
  title={Zero-Knowledge Proof Frameworks: A Systematic Survey},
  author={Sheybani, Nojan and Ahmed, Anees and Kinsy, Michel and Koushanfar, Farinaz},
  journal={arXiv preprint arXiv:2502.07063},
  year={2025}
}

@article{scaramuzza2025engineering,
  title={Engineering Trustworthy Machine-Learning Operations with Zero-Knowledge Proofs},
  author={Scaramuzza, Filippo and Quattrocchi, Giovanni and Tamburri, Damian A},
  journal={arXiv preprint arXiv:2505.20136},
  year={2025}
}

@article{xing2025zero,
  title={Zero-Knowledge Proof-Based Verifiable Decentralized Machine Learning in Communication Network: A Comprehensive Survey},
  author={Xing, Zhibo and Zhang, Zijian and Zhang, Ziang and Li, Zhen and Li, Meng and Liu, Jiamou and Zhang, Zongyang and Zhao, Yi and Sun, Qi and Zhu, Liehuang and others},
  journal={IEEE Communications Surveys \& Tutorials},
  year={2025},
  publisher={IEEE}
}

@article{fan2024psvcnn,
  title={psvcnn: A zero-knowledge cnn prediction integrity verification strategy},
  author={Fan, Yongkai and Xu, Binyuan and Zhang, Linlin and Tan, Gang and Yu, Shui and Li, Kuan-Ching and Zomaya, Albert},
  journal={IEEE Transactions on Cloud Computing},
  volume={12},
  number={2},
  pages={359--369},
  year={2024},
  publisher={IEEE}
}

@inproceedings{zhang2020dynamic,
  title={Dynamic slicing for deep neural networks},
  author={Zhang, Ziqi and Li, Yuanchun and Guo, Yao and Chen, Xiangqun and Liu, Yunxin},
  booktitle={Proceedings of the 28th ACM Joint Meeting on European Software Engineering Conference and Symposium on the Foundations of Software Engineering},
  pages={838--850},
  year={2020}
}

@article{zhou2024neusemslice,
  title={NeuSemSlice: Towards Effective DNN Model Maintenance via Neuron-level Semantic Slicing},
  author={Zhou, Shide and Li, Tianlin and Huang, Yihao and Shi, Ling and Wang, Kailong and Liu, Yang and Wang, Haoyu},
  journal={ACM Transactions on Software Engineering and Methodology},
  year={2024},
  publisher={ACM New York, NY}
}

@misc{ezkl2025,
  author       = {{Zkonduit Inc.}},
  title        = {{EZKL}: The {EZKL} System},
  howpublished = {\url{https://docs.ezkl.xyz/}},
  note         = {Accessed: \today},
%  year         = {2025}
}

@misc{zkmlblueprints2025,
  author       = {{Inference Labs Inc.}},
  title        = {zkML Blueprints: Arithmetic Circuits for Neural Network Inference},
  howpublished = {\url{https://github.com/inference-labs-inc/zkml-blueprints}},
  note         = {Accessed: \today},
%  year         = {2025}
}

@misc{expandercc2025,
  author       = {{Polyhedra Network}},
  title        = {{E}xpander {C}ompiler {C}ollection},
  howpublished = {\url{https://github.com/PolyhedraZK/ExpanderCompilerCollection}},
  note         = {Accessed: \today},
%  year         = {2025}
}

@misc{lenethf2025,
  author       = {{MindSpore AI}},
  title        = {{MindSpore Image Classification Models with MNIST on the Hugging Face Hub}},
  howpublished = {\url{https://huggingface.co/mindspore-ai/LeNet}},
  note         = {Accessed: \today},
%  year         = {2025}
}

@misc{lenetkaggle2025,
  author       = {Paras Varshney},
  title        = {{LeNet Architecture: A Complete Guide}},
  howpublished = {\url{https://www.kaggle.com/code/blurredmachine/lenet-architecture-a-complete-guide}},
  note         = {Accessed: \today},
%  year         = {2020}
}

@ARTICLE{lecun1998gradient,
  author={Lecun, Y. and Bottou, L. and Bengio, Y. and Haffner, P.},
  journal={Proceedings of the IEEE}, 
  title={Gradient-based learning applied to document recognition}, 
  year={1998},
  volume={86},
  number={11},
  pages={2278-2324},
  doi={10.1109/5.726791}}

\end{document}